\documentclass[conference]{IEEEtran}
\IEEEoverridecommandlockouts
\usepackage{cite}
\usepackage{amsmath,amssymb,amsfonts}
\usepackage{algorithmic}
\usepackage{graphicx}
\usepackage{textcomp}
\usepackage{xcolor}
\usepackage{booktabs}
\usepackage{caption}
\usepackage[hidelinks]{hyperref}

\begin{document}

\title{Distill3R: A Pipeline for Democratizing 3D Foundation Models on Commodity Hardware}

\author{
    \IEEEauthorblockN{Brandon Leblanc}
    \IEEEauthorblockA{\textit{Immersive and Creative Technologies Lab} \\
    \textit{Concordia University}\\
    Montreal, Canada \\
    brandon.leblanc@mail.concordia.ca}
    \and
    \IEEEauthorblockN{Charalambos Poullis}
    \IEEEauthorblockA{\textit{Immersive and Creative Technologies Lab} \\
    \textit{Concordia University}\\
    Montreal, Canada \\
    charalambos.poullis@concordia.ca}
}

\maketitle


\begin{abstract} While multi-view 3D reconstruction has shifted toward large-scale foundation models capable of inferring globally consistent geometry, their reliance on massive computational clusters for training has created a significant barrier to entry for most academic laboratories. To bridge this compute divide, we introduce Distill3R, a framework designed to distill the geometric reasoning of 3D foundation models into compact students fully trainable on a single workstation. Our methodology centers on two primary innovations: (1) an offline caching pipeline that decouples heavy teacher inference from the training loop through compressed supervision signals, and (2) a confidence-aware distillation loss that leverages teacher uncertainty to enable training on commodity hardware. We propose a 72M-parameter student model which achieves a 9x reduction in parameters and a 5x inference speedup compared to its 650M-parameter teacher. The student is fully trainable in under 3 days on a single workstation, whereas its teacher requires massive GPU clusters for up to a week. We demonstrate that the student preserves the structural consistency and qualitative geometric understanding required for functional 3D awareness. By providing a reproducible, single-workstation training recipe, Distill3R serves as an exploratory entry point for democratized 3D vision research and efficient edge deployment. This work is not intended to compete with state-of-the-art foundation models, but to provide an accessible research baseline for laboratories without access to large-scale compute to train and specialize models on their own domain-specific data at minimal cost. \end{abstract}

\begin{center}
\small
Source code available at \href{https://github.com/TheFourthKaramazov/Distill3R}{https://github.com/TheFourthKaramazov/Distill3R}
\end{center}


\section{Introduction}
\label{sec:intro}

Unconstrained multi-view reconstruction methods that directly predict 3D geometry in a single forward pass have advanced rapidly since the introduction of DUSt3R \cite{wang2024dust3r}. By demonstrating that a neural network could infer consistent geometry and implicit relative camera poses from image pairs without following the classical pipeline of Structure-from-Motion (SfM) and Multi-View Stereo (MVS) \cite{cui2017hsfm, wei2020deepsfm, schonberger2016sfm, agarwal2009building, hartley2004multiple, snavely2006photo, ozyesil2017survey}, DUSt3R marked a turning point in learned 3D reconstruction. Its pairwise reconstruction and subsequent procedural global alignment strategy have been surpassed by fully end-to-end architectures capable of processing all views jointly and producing globally consistent reconstructions in a single step \cite{yang2025fast3r, wang2025vggt, leroy2024mast3r, zhang2025flare, zhang2025monst3r, tang2024mvdust3r, duisterhof2025mast3r}. Using global attention mechanisms and large-scale training, these models follow a familiar trajectory in deep learning: rapid gains driven by exponential increases in both model capacity and dataset scale \cite{brown2020language, dosovitskiy2021image, simeoni2025dinov3, achiam2023gpt4, radford2021learning, kaplan2020scaling, hoffmann2022training, peebles2023scalable, wang2026pi3}. For example, Fast3R \cite{yang2025fast3r} is trained on 128 A100 GPUs for 6.13 days and VGGT \cite{wang2025vggt} on 64 A100 GPUs for 9 days, containing approximately $650~\mathrm{M}$ and $1~\mathrm{B}$ parameters, respectively. Although these advances have shown impressive results, they have also concentrated progress within a small number of frontier labs equipped with vast computational resources, thereby preventing most labs from building upon and improving these models, which results in a bottleneck for research progress in this domain.  Moreover, their size and inference cost make them impractical for deployment on edge devices, where 3D reconstruction is most valuable, such as in robotics, medical applications, and autonomous navigation \cite{wu2022tinyvit, xia2024rgbd,baruch2021arkitscenes, pan2023aria,szot2021habitat, sun2025avggtrethinkingglobalattention}.

This disparity between high-capacity research models and real-world deployment environments has created a divergence between innovation and accessibility. While breakthroughs in multi-view geometry increasingly rely on massive computation, many research and industry groups lack the resources to train or deploy such systems. This is evidenced by recent work surrounding VGGT being training-free \cite{sun2025avggtrethinkingglobalattention, wang2025fastervggtblocksparseglobal, fastvggt_arxiv, quantvggt_arxiv}, demonstrating an inherent bias towards improving efficiency and performance while avoiding training. As a result, the community faces a structural bottleneck: the pursuit of absolute performance has outpaced our collective ability to participate in the development and iteration of these models, gate-keeping research innovation. Addressing this requires a paradigm shift that prioritizes the establishment of accessible research pipelines over marginal gains in accuracy. Rather than attempting to replicate the industrial-scale infrastructure of frontier labs, the primary contribution of this work is to provide a high-efficiency methodology that lowers the barrier to entry. By establishing a reproducible framework for distillation, we enable the broader community to leverage high-capacity models by proxy and explore new research directions in 3D vision using commodity hardware.

To this end, we introduce Distill3R, a novel distillation framework \cite{hinton2015distillation, gou2020knowledge, touvron2021training, wu2022tinyvit} designed to address this accessibility problem in 3D vision. Distill3R enables researchers and practitioners to train compact, high-performance reconstruction models on a single workstation, achieving up to a $9\times$ reduction in model size relative to the teacher. We adopt Fast3R \cite{yang2025fast3r} as the teacher due to its balance of efficiency and accuracy. Inspired by TinyViT \cite{wu2022tinyvit}, Distill3R introduces an offline cache mechanism that stores teacher-generated supervision signals, eliminating the need for expensive online teacher inference during training. By transforming foundation model distillation into a task feasible on commodity hardware, this approach enables scalable experimentation and rapid iteration without the need for multi-million dollar infrastructure.

Through our evaluation, we show that a $72~\mathrm{M}$-parameter student trained via Distill3R serves as a functional, exploratory entry point for resource-constrained 3D vision research. While absolute  precision on benchmarks like 7-Scenes exhibits a quantitative drop compared to the teacher, we demonstrate that the student successfully preserves the structural and topological consistency required for general spatial awareness and monocular 3D reconstruction as well as a significantly superior adherence to absolute metric scale across both indoor and extreme out-of-distribution (OOD) object-centric scenes. This consistency is crucial for downstream robotics tasks such as reactive obstacle avoidance and local occupancy mapping \cite{hornung2013octomap}, where high update rates and qualitative scene understanding often outweigh millimetric precision \cite{davison2007monoslam, teed2021droid, qin2018vins}. We identify that the observed quantitative gap is primarily driven by subtle geometric warping and scale ambiguities that impact regression metrics but do not preclude functional utility. Despite these trade-offs, our confidence-aware distillation loss successfully stabilizes training on commodity hardware, yielding a student with a $5\times$ inference speedup. By shifting the focus from inference-only deployment to an accessible training pipeline, we empower the community to modify and refine these models without requiring industrial-scale infrastructure.

Our main technical contributions are as follows:

1. Distill3R, a novel distillation framework for learned 3D reconstruction that employs an offline cache mechanism to enable training on a single workstation.

2. A confidence-aware distillation loss, which leverages cached teacher confidence values for both direct supervision and geometric loss weighting, improving training stability by taking advantage of learned teacher confidence values for geometric weighting rather than degenerate student predictions.

3. Democratized research baseline: We provide a comprehensive characterization of the trade-offs between model size and reconstruction fidelity, offering a $9\times$ parameter reduction and $5\times$ inference speedup as an exploratory roadmap for accessible 3D vision research.
\section{Related Work}
\label{sec:related}

\subsection{Procedural and differentiable pipelines}
The traditional Structure-from-Motion (SfM) pipeline \cite{schonberger2016sfm, hartley2004multiple, snavely2006photo}, composed of sequential steps that include feature extraction \cite{detone2018superpoint, dusmanu2019d2net, zhao2023aliked, yi2016lift, lowe2004distinctive}, feature matching \cite{sarlin2020superglue, sun2021loftr, edstedt2024roma, lindenberger2023lightglue_iccv}, correspondence search, sparse reconstruction, and multi-view stereo (MVS) \cite{yao2018mvsnet, gu2020cascade, cheng2022masked}, remains the most accessible and interpretable approach to recover 3D geometry from unordered images when hardware resources are limited. However, each stage of this pipeline introduces cumulative error and often fails under real-world conditions such as textureless regions, limited overlap, or dynamic scenes. Furthermore, the entire process can take hours as the number of input images increases \cite{schonberger2016sfm, wu2013towards}. Although recent work has integrated learning-based improvements such as robust correspondence search \cite{sarlin2020superglue, edstedt2023dkm}, learned features \cite{tyszkiewicz2020disk, caron2021emerging}, and differentiable bundle adjustment \cite{wang2024vggsfm, tang2018ba, wei2020deepsfm, teed2018deepv2d}, the pipeline remains slow, fragile, and cumbersome to deploy outside controlled datasets.

\subsection{Geometric foundation models}
To address the limitations of procedural pipelines, the field turned toward single pass, feed-forward networks. DUSt3R pioneered this direction by regressing dense point maps from image pairs. The two important contributions that make this possible are carefully curated datasets \cite{reizenstein2021co3d, dai2017scannet, zheng2023pointodyssey, deitke2023objaverse, baruch2021arkitscenes, szot2021habitat, li2018megadepth, yao2020blendedmvs, yeshwanth2023scannetpp} and the point map representation \cite{wang2024dust3r}. Rather than solving for camera intrinsics and extrinsics explicitly to triangulate points \cite{hartley2004multiple}, DUSt3R and subsequent models \cite{cabon2025must3r, jang2025pow3r, leroy2024mast3r, wang2025vggt, yang2025fast3r, zhang2025monst3r, wang2025spann3r, wang2026pi3} train a model to directly regress a dense 2D coordinate map with the same dimensions as the input images, creating a one-to-one mapping between image pixels and 3D coordinates. This representation unifies geometry and pose while allowing geometric priors to be learned through the mapping from pixel to 3D point. These models also predict a pixel-wise confidence map with the same dimensions as the input image and the output point map, allowing the system to learn its own uncertainty and filter out unreliable predictions. This dense and confidence-aware regression serves as the foundation of our objective function in the distillation framework. However, these methods \cite{wang2024dust3r, leroy2024mast3r, duisterhof2025mast3r} are fundamentally pairwise; achieving global consistency across a scene requires costly post-hoc global alignment solvers such as PnP \cite{lepetit2009epnp, wang2023posediffusion, kendall2015posenet} that scale poorly. Sequential approaches like \cite{wang2025spann3r, zhang2025monst3r} attempt to mitigate this via memory mechanisms but suffer from drift accumulation over long sequences. Most recently, global foundation models like Fast3R~\cite{yang2025fast3r}, VGGT~\cite{wang2025vggt}, and $\mathrm{\pi}^3$\cite{wang2026pi3} have achieved state-of-the-art results by processing all views simultaneously using global attention mechanisms. Although these models solve the global consistency problem in a single forward pass, they do so at a significant computational cost. With parameters ranging from  $650~\mathrm{M}$ to more than  $1~\mathrm{B}$ and training requirements exceeding 100 A100 GPUs, these heavyweight models remain inaccessible for research iteration and deployment on edge devices, reinforcing the infrastructure bottleneck that prevents the broader community from participating in the development of 3D vision foundation models.

\subsection{Efficient 3D vision and distillation}
Knowledge Distillation (KD)~\cite{hinton2015distillation, gou2020knowledge, touvron2021training} is a standard technique for model compression, yet its application to generalizable 3D reconstruction remains underexplored. Distilled encoders such as DUNE \cite{sariyildiz2025dune}, which we use in our student model, have managed to surpass their teachers' performance in 3D tasks by using heterogeneous co-distillation to achieve knowledge synthesis and refinement while significantly reducing parameter count. Previous efforts in 3D-aware KD have largely focused on disparate tasks, such as compressing NeRFs \cite{wei2021nerfingmvs, tang2024lgm} or depth estimation \cite{yang2024depthv2, wu2023adudepth, song2023msdpt}. In the multi-view domain, KD-MVS \cite{ding2022kdmvs} applies distillation to improve depth accuracy without ground-truth labels, but it does not prioritize model compression or reducing heavy memory requirements and relies on the restrictive assumption of known camera parameters. Close to our work, \cite{dutt2024multiview} distills DUSt3R but remains limited to per-scene optimization which is of limited use. They do not attempt to distill a general-purpose model. Concurrently, inference-only acceleration methods have emerged: QuantVGGT \cite{quantvggt_arxiv} applies post-training quantization to compress VGGT to 4 bits; AVGGT \cite{sun2025avggtrethinkingglobalattention} uses mechanistic interpretability insights \cite{stary2025understandingmultiviewtransformers} to remove global attention in early decoder layers where correspondences are not optimized; Faster VGGT \cite{wang2025fastervggtblocksparseglobal} uses block sparse attention to increase efficiency; while FastVGGT \cite{fastvggt_arxiv} utilizes token pruning to speed up inference. Although effective for deployment, these are strictly inference-time optimizations that do not address the high computational requirements of training. This creates a documented bias in current research toward improving performance while avoiding the training process due to prohibitive infrastructure costs. 
\section{Method}
\label{sec:method}

\subsection{Problem Definition}
\label{sec:problem-def}
We address the task of unconstrained multi-view 3D reconstruction. Given a set of $N$ unposed 2D images $\mathcal{I} = \{I_1, \dots, I_N\}$ depicting a scene, the goal is to infer the 3D geometry of the scene. Let $T$ be a pre-trained teacher model and $S$ be our compact student model. For any given image $I_k \in \mathcal{I}$, both models are trained to predict dense 2D maps that encode 3D information at a resolution $H \times W$. The specific outputs we aim to distill from the teacher $T$ are:

\begin{itemize}
    \item Global 3D point map: $\mathbf{P}^{g}_k \in \mathbb{R}^{H \times W \times 3}$. For each pixel $(i,j)$ in image $I_k$, this map stores the predicted 3D coordinate $\mathbf{p}^{g}_{k, ij}$ in a single, consistent global reference frame for the entire scene.
    \vspace{0.1cm}
    \item Local 3D point map: $\mathbf{P}^{\ell}_k \in \mathbb{R}^{H \times W \times 3}$. For each pixel $(i,j)$, this map stores the 3D coordinate $\mathbf{p}^{\ell}_{k, ij}$ relative to its own local camera frame. 
    \vspace{0.1cm}
    \item Confidence maps: $C^{g}_k, C^{\ell}_k \in \mathbb{R}^{H \times W}$. These maps store the teacher's per-pixel confidence $c_{k, ij} \in [0, \infty)$ in the accuracy of its corresponding global and local 3D predictions.
\end{itemize}

Our objective is to train the student model $S$ such that its predicted outputs, denoted $(\mathbf{P}^{s,g}, \mathbf{P}^{s,\ell}, C^{s,g}, C^{s,\ell})$, match the teacher's outputs $(\mathbf{P}^{t,g}, \allowbreak \mathbf{P}^{t,\ell}, \allowbreak C^{t,g}, \allowbreak C^{t,\ell})$. This is achieved by optimizing $S$ with a specialized distillation loss $\mathcal{L}_{\text{total}}$, detailed in Section \ref{sec:loss}.

\begin{figure*}[t!]
    \centering
    \includegraphics[width=1.0\textwidth]{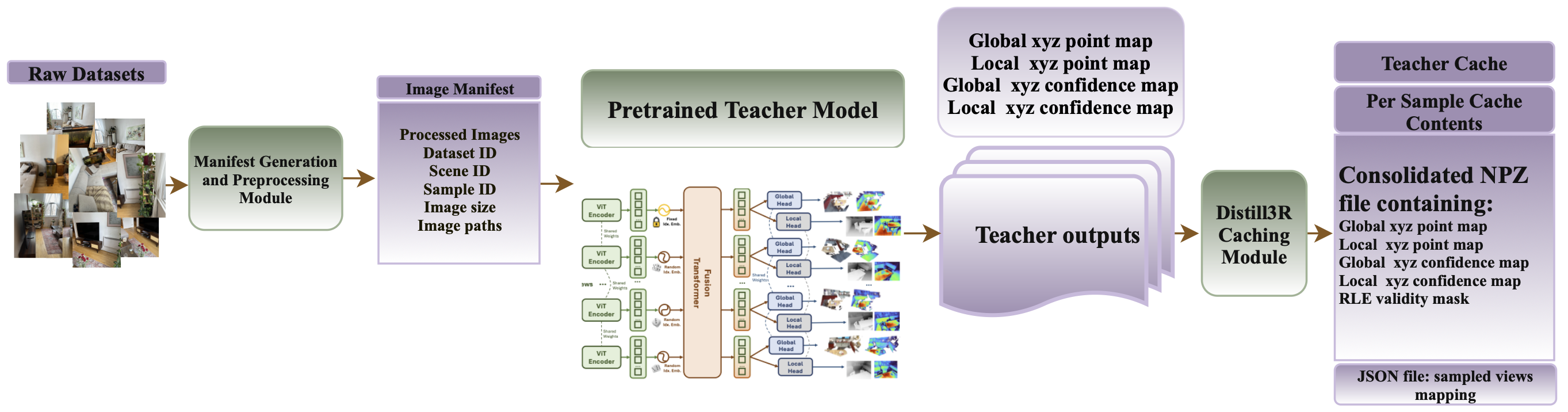}
    \caption{An overview of the offline teacher caching pipeline.
    We decouple training by pre-computing teacher supervision. A Manifest Generation module indexes raw data, which is processed by Fast3R in fixed-size samples. The caching module compresses predictions (\texttt{float16}) and masks (RLE) into a compressed static cache used by the student and teacher dataset modules for training.}
    \label{fig:cache_pipeline}
\end{figure*}

\subsection{Offline Teacher Caching}
\label{sec:cache}
To enable training on a single workstation, we implement a two-stage offline caching pipeline (Fig. \ref{fig:cache_pipeline}). By pre-computing teacher predictions, we decouple the heavy inference cost from the student training loop, transforming the distillation task into an accessible supervised learning problem that establishes a reproducible training recipe for labs with limited compute.

\subsubsection{Data Unification and Pre-processing}
Raw datasets are processed with a manifest generation module that performs two tasks: metadata standardization and image processing. This module processes the source images into the expected format and generates a complete \textit{Image Manifest}.

This manifest logs normalized metadata for every processed frame, specifically: the \textit{dataset ID}, \textit{scene ID}, \textit{sample ID}, \textit{image dimensions}, and \textit{absolute paths}. The \textit{scene ID} groups all $N$ views of a sequence, enabling retrieval of geometrically consistent batches. The processed image corpus and the unified manifest serve as shared input components for both offline cache generation and online student training.

\subsubsection{Cache Generation and Export}
For each sample in the manifest, the pipeline loads the corresponding images and passes them as a joint batch to the teacher. This single forward pass generates the full set of four supervision maps $(\mathbf{P}^{t,g}, \allowbreak \mathbf{P}^{t,\ell}, \allowbreak C^{t,g}, \allowbreak C^{t,\ell})$ as formally defined in Sec. \ref{sec:problem-def}.

These raw, high-precision outputs then undergo processing to optimize storage efficiency and I/O throughput:

\begin{enumerate}
    \item \textit{Resolution alignment.} The raw teacher outputs are down-sampled to a target resolution of $224 \times 518$. This resolution is deliberately chosen to be divisible by the $14 \times 14$ patch size of the student's DUNE encoder \cite{sariyildiz2025dune}. Pre-aligning resolutions eliminates the need for on-the-fly resizing, preventing aliasing artifacts and removing interpolation overhead that can overload CPUs during the training loop.
    
    \item \textit{Supervision filtering.} We filter predictions using the teacher's own uncertainty. We apply a confidence threshold $\mathrm{\tau}=0.3$ to the local confidence map $C^{t,\ell}$, following empirical guidelines for point map regression models \cite{wang2024dust3r, tang2024mvdust3r}, to generate a binary \textit{validity mask} $\mathcal{M}$. Pixels where $c^{t,\ell}_{ij} < \tau$ are explicitly masked and ignored by the loss function (Sec. \ref{sec:loss}), preventing the student from learning teacher errors and decreasing computation overhead during training.
    
    \item \textit{Compression.} Given the $7\times$ storage expansion from raw data to cache, we apply aggressive compression. Continuous maps ($\mathbf{P}, C$) are quantized from \texttt{float32} to \texttt{float16}, while the sparse validity masks $\mathcal{M}$ are compressed via Run-Length Encoding (RLE). This reduces the data expansion from $7\times$ to $4\times$, ensuring that data throughput remains high enough to saturate the GPU and minimize idle time during the student training loop.
    
    \item \textit{Serialization and stacking.} The processed data for all $N$ views are aggregated. Instead of saving individual files, we stack the $N$ maps along the batch dimension and save them into a single consolidated archive. This consolidation is a necessary I/O optimization: it reduces file system interactions by orders of magnitude, preventing file-handle constraints and ensuring data loading keeps pace with the GPU.

\end{enumerate}

\begin{figure}[t]
    \centering
    \includegraphics[width=1.0\columnwidth]{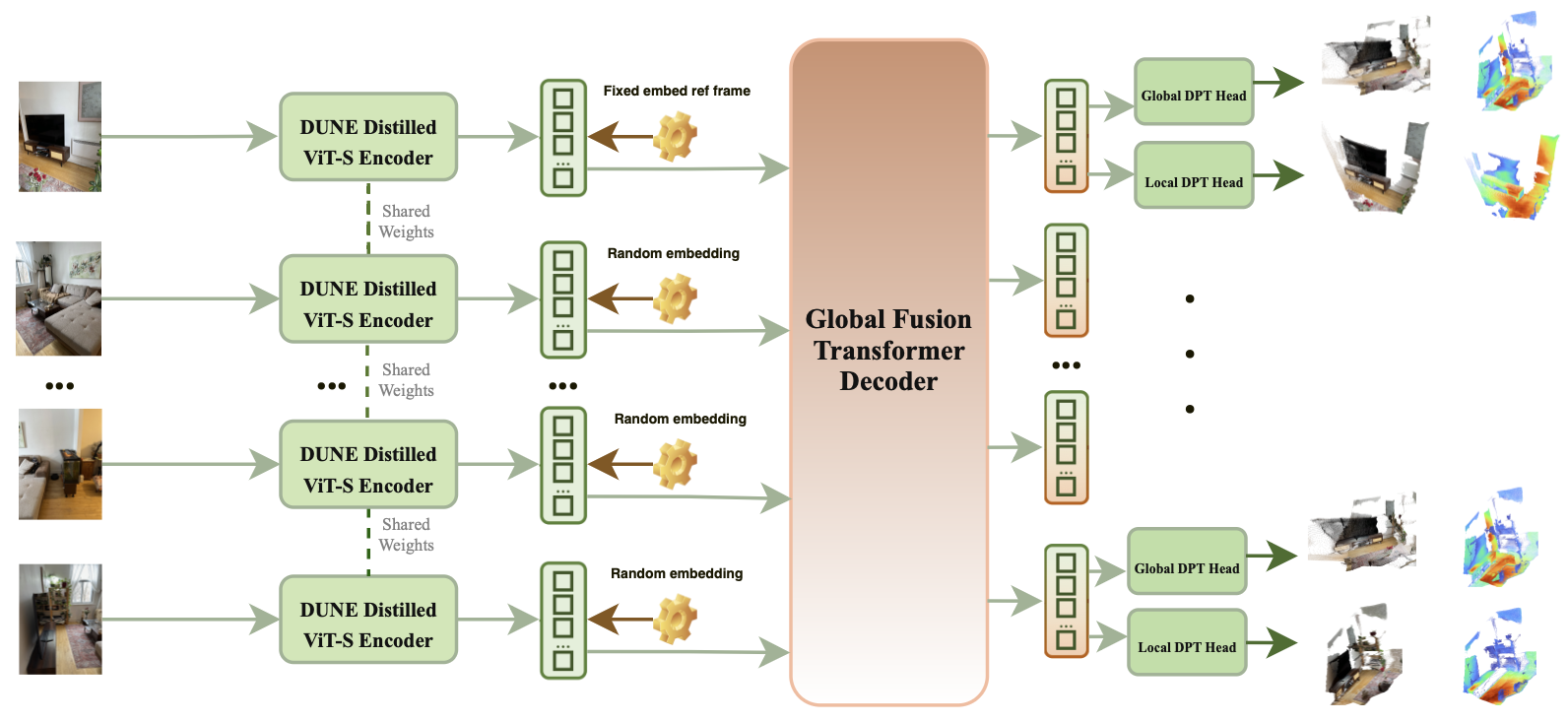}
    \caption{Student Architecture. DUNE ViT-S encoder followed by a 6L/384d Global Fusion Transformer and DPT heads.}
    \label{fig:student_arch}
\end{figure} 

\subsection{Student Model Architecture}
\label{sec:architecture}

We design the student $S$ to mirror the teacher's global reasoning capabilities while fitting within consumer hardware constraints and significantly reducing the parameter count. As shown in Fig. \ref{fig:student_arch}, we retain the meta-architecture of the teacher, but replace heavy components with lightweight alternatives.

\textit{Encoder.} Instead of the teacher's heavy CroCo \cite{weinzaepfel2022croco} backbone, we employ \textit{DUNE ViT-S} \cite{sariyildiz2025dune} with weights shared across all $N$ views. DUNE is selected for its robust 3D-aware feature space and small parameter count that outperforms larger baselines.

\textit{Embeddings.} Input images are divided into non-overlapping 14x14 patches. Following Fast3R, learned positional embeddings are added to these patch tokens to encode spatial information. The first view is set as the reference frame by default. We add a fixed embedding to the reference frame because it represents the origin in the global coordinate frame. Randomized view ID embeddings are used for subsequent views, enabling the model to distinguish between different viewpoints while providing implicit data augmentation through randomized assignment at each forward pass.

\textit{Global fusion decoder and dense prediction heads.} We retain the global self-attention mechanism to resolve scale and alignment implicitly across views but we significantly reduce the depth and width relative to the teacher. This fused representation is then decoded by two lightweight DPT heads \cite{ranftl2021vision} to predict global ($\mathbf{P}^{s,g}$) and local ($\mathbf{P}^{s,\ell}$) coordinates along with their confidences.

\subsection{Distillation loss function}
\label{sec:loss}
Our total distillation loss, $\mathcal{L}_{\text{total}}$, is designed to transfer both geometric accuracy and implicitly learned uncertainty from the teacher to the student. It is a weighted sum of three loss components: a loss for the global 3D point map, a loss for the local 3D point map, and a loss for the local and global confidence maps.

All loss components are computed only on the set of valid pixels $\mathcal{M}$, as defined by the validity mask (Section \ref{sec:cache}). Let $|\mathcal{M}|$ be the total number of valid pixels in a batch. The total loss is defined as:
\begin{equation}
\label{eq:total_loss}
\mathcal{L}_{\text{total}} = \mathcal{L}_{\text{geom\_total}} + \gamma \mathcal{L}_{\text{conf}},
\end{equation}
where $\alpha_g$, $\alpha_\ell$, and $\gamma$ are hyperparameters balancing the loss terms. Based on our experiments, the gradient magnitudes for the confidence loss require down-weighting and the global geometry loss requires an increased weight: $\alpha_g = 2.0$, $\alpha_\ell = 1.0$, and $\gamma = 0.001$.

\subsubsection{Confidence-aware geometric loss}
The primary objective is to regress the teacher's dense geometry. We define the geometric loss for both global ($g$) and local ($\ell$) heads as a weighted MSE:
\begin{equation}
\label{eq:geom_loss}
\mathcal{L}_{\text{g}} = \frac{1}{|\mathcal{M}|} \sum_{k \in \mathcal{I}} \sum_{(i,j) \in \mathcal{M}_k} w^{m}_{k,ij} \| \mathbf{p}^{s,m}_{k,ij} - \mathbf{p}^{t,m}_{k,ij} \|_2^2,
\end{equation}
where $m \in \{g, \ell\}$ denotes the global or local coordinate map being regressed and $w^m_{k,ij}$ denotes the cached teacher confidence. The total geometric contribution to the final objective is defined as:
\begin{equation}
\mathcal{L}_{\text{geom\_total}} = \alpha_g \mathcal{L}_{g} + \alpha_\ell \mathcal{L}_{\ell}.
\end{equation}
A primary contribution of this loss is the design of the weighting term $w^m_{k,ij}$. We explicitly derive this weight from the teacher's cached confidence $C^t$, rather than the student's predicted confidence $C^s$. In standard uncertainty-aware regression, models often learn to predict low confidence ($C^s \to 0$) to attenuate the loss in difficult regions \cite{kendall2017what, wang2024dust3r}, leading to degenerate solutions early in training. By setting the loss weighting to the fixed calibrated teacher confidence, we prevent this collapse \cite{wu2023adudepth, kendall2017what}, providing a stable optimization landscape that allows for successful convergence even within the shorter training cycles necessitated by resource-constrained environments. The teacher's confidence acts as a stable signal, forcing the student to prioritize high-certainty geometric areas while down-weighting uncertain regions.

\textit{Scale-invariant normalization.} To enforce geometric learning, distinct normalization strategies are used. For the global loss $\mathcal{L}_{g}$, we normalize all views using a shared scale factor $s$ computed across the entire batch:
\begin{equation}
   \tilde{\mathbf{p}}^{g}_{k,ij} = \frac{\mathbf{p}^{g}_{k,ij}}{s}, \quad 
s = \frac{1}{|\mathcal{M}|} \sum_{k \in \mathcal{I}} \sum_{(i,j) \in \mathcal{M}_k} \| \mathbf{p}^{g}_{k,ij} \|_2.
\end{equation}
This shared scalar enforces a consistent global coordinate frame. Conversely, for the local loss $\mathcal{L}_{\ell}$, each view is normalized independently by its own view-specific scale factor $s_k$:
\begin{equation}
    \tilde{\mathbf{p}}^{\ell}_{k,ij} = \frac{\mathbf{p}^{\ell}_{k,ij}}{s_k}, \quad 
    s_k = \frac{1}{|\mathcal{M}_k|} \sum_{(i,j) \in \mathcal{M}_k} \| \mathbf{p}^{\ell}_{k,ij} \|_2.
\end{equation}
This ensures the model generalizes to arbitrary view combinations rather than overfitting to the absolute scale of training scenes.

\subsubsection{Confidence distillation loss}
We directly distill the teacher's confidence maps. This teaches the student to reproduce the teacher's uncertainty calibration, which is important for geometric reliability and for downstream applications that use this confidence signal \cite{kendall2017what}. The student's confidence outputs ($C^{s,g}, C^{s,\ell}$) are supervised using a mean L1 loss against the cached teacher confidences ($C^{t,g}, C^{t,\ell}$) over the valid pixel mask $\mathcal{M}$:

\begin{equation}
\label{eq:conf_loss}
\mathcal{L}_{\text{conf}} =
\frac{1}{2|\mathcal{M}|}
\sum_{k \in \mathcal{I}} \sum_{(i,j) \in \mathcal{M}_k}
\bigl(
| c^{s,g}_{k,ij} - c^{t,g}_{k,ij} |
+ | c^{s,\ell}_{k,ij} - c^{t,\ell}_{k,ij} |
\bigr).
\end{equation}

This loss, combined with geometric losses, encourages the student model to learn a complete representation of the 3D scene.
\section{Experiments}
\label{sec:experiments}

\begin{figure*}[t!]
    \centering
    \includegraphics[width=1.0\textwidth]{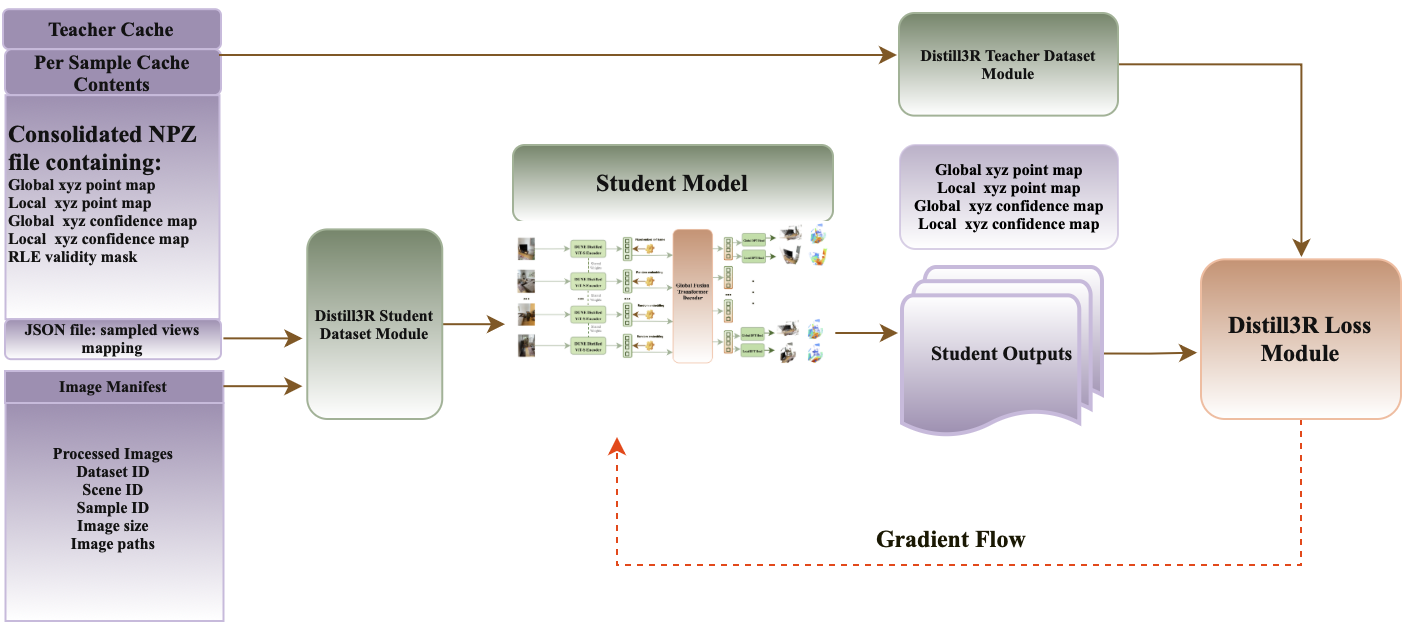}
    \caption{An overview of the Distill3R training loop.
    (1) The Distill3R Student Dataset Module loads a batch of $N$ images (from the Image Manifest) and the corresponding $N$ pre-computed supervision signals (from the Teacher Cache).
    (2) The Student Model processes the images and generates its own 3D point maps and confidence maps.
    (3) The Distill3R Loss Module compares the Student Outputs against the cached Teacher Outputs to compute $\mathcal{L}_{\text{total}}$.
    (4) The resulting Gradient Flow (red dotted line) updates only the Student Model's parameters.}
    \label{fig:training_pipeline}
\end{figure*}

\textit{Compute setup.}
All models are trained on a single workstation equipped with two NVIDIA RTX 6000 Ada (48~GB) GPUs. The full training pipeline is also verified to run on a single GPU without modification. Unless otherwise stated, all experiments and evaluations are conducted using a single RTX 6000 Ada GPU.

\textit{Datasets.}
 The training set is constructed from six established datasets following Fast3R: CO3D-v2 \cite{reizenstein2021co3d}, ScanNet++ \cite{yeshwanth2023scannetpp}, Habitat \cite{savva2019habitat}, MegaDepth \cite{li2018megadepth}, BlendedMVS \cite{yao2020blendedmvs}, and ARKitScenes \cite{baruch2021arkitscenes}. This mix ensures that the student sees a wide distribution of scene types, scene scales, lighting conditions, and sensor characteristics, ranging from object-centric captures, indoor navigation, to large-scale outdoor scenes.

\textit{Dynamic sub-sampling strategy.} 
\label{sec:dynamic}
Training on all available frames is computationally redundant due to the high overlap between frames which makes frame to frame variation indistinguishable in most scenes. To maximize geometric variance while minimizing dataset size, a dynamic sub-sampling strategy is used. The sampling stride is adjusted based on scene type, overlap, and number of images. For example, in object-centric data (CO3D \cite{reizenstein2021co3d}), a wide baseline dependent on the number of images is used to encourage global understanding; for navigation datasets (Habitat \cite{savva2019habitat}, ScanNet++ \cite{yeshwanth2023scannetpp}), the stride is increased proportional to trajectory length to ensure sufficient overlap for feature matching; for large outdoor scenes (MegaDepth \cite{li2018megadepth}), the stride is increased based on scene size assuming significant overlap and redundant information. We sample $N=20$ contiguous frames following \cite{yang2025fast3r}. This strict temporal locality preserves the consistency required for the student's global fusion mechanism to learn cross-view correspondences \cite{stary2025understandingmultiviewtransformers}. This process distills the multi-terabyte corpus down to a dense training set of approximately 450k images (150~GB).

\textit{Student Architecture.} 
Our $72~\mathrm{M}$-parameter student model consists of three key components:

\begin{enumerate}
    \setlength\itemsep{0em}
    \item \textit{Image Encoder.} DUNE ViT-Small (21M) initialized with DUNE pretrained weights. It operates on $14 \times 14$ patches with weights shared across all $N$ views.
    \item \textit{Fusion Transformer.} We replace the teacher's heavy decoder with a compressed transformer (6 layers, 6 heads, 384 embedding dim) initialized from scratch. It uses cross view self-attention to fuse features across the view sequence.
    \item \textit{Pointmap Heads.} Two lightweight DPT heads regress dense global/local coordinates and confidence maps.
\end{enumerate}

\textit{Training Pipeline.} To make training possible on a single workstation, the training loop assumes that the cache has already been generated. As a result, the training process is similar to standard supervised training. The cache and manifest represent the dataset and are processed using the student dataset module and teacher dataset module as depicted in Fig. \ref{fig:training_pipeline}. The student dataset module retrieves image batches and the associated teacher-sampled paths for those images from the manifest. In parallel, the teacher dataset module fetches pre-computed supervision signals (geometry and confidence) directly from the consolidated cache corresponding to paths processed by the student dataset module for that batch. The stacking step in cache processing ensures that we do not reach file handle limits and overload the CPU cores on a single workstation by aggregating all the supervision signals for each sample into single files. The cache generation is completed on a single NVIDIA RTX 6000 Ada and takes 11.3~hours. 

\textit{Training Hyperparameters.}
Training utilizes the AdamW optimizer \cite{loshchilov2017decoupled} with a weight decay of $0.01$. We employ a cosine annealing schedule \cite{loshchilov2016sgdr} with linear warmup for the first 3,500 steps, peaking at $1 \times 10^{-4}$ before decaying to $5 \times 10^{-5}$. To maximize throughput, the model exploits \texttt{bf16} mixed-precision \cite{micikevicius2017mixed} and FlashAttention \cite{dao2022flashattention, dao2024flashattention2}, enabling scaling to $N=20$ views without memory exhaustion. All experiments are conducted on a single workstation with 2$\times$ NVIDIA RTX 6000 Ada (48~GB) GPUs for a total of 60 epochs over 2.875~days using a resolution of $224\times 518$. Using a per-GPU batch size of 4 with 2 gradient accumulation steps yields an effective batch size of 16. 

\subsection{Inference Efficiency}

The motivation for Distill3R is to retain the reconstruction capabilities of large feed-forward models while drastically reducing computational overhead. We measure inference latency and peak GPU memory usage on a single NVIDIA RTX 6000 Ada Generation GPU (48~GB).

We adopt a sliding-window protocol relevant to real-time applications, scaling the number of input views $N \in \{12, 32, 64, 96, 128\}$ at a fixed resolution of $378\times 518$ for all models except Fast3R. Fast3R uses a fixed resolution of $384\times 512$ to match CroCo patch size.  We measure the wall-clock time required to process a single batch and track the peak VRAM usage during the forward pass. To ensure a fair comparison, we only test single pass feed-forward models that utilize global attention because models like DUSt3R that require post-processing for alignment are notably slower and more memory intensive \cite{yang2025fast3r}. We remove the maximum parallel views restriction for the dense predictions heads as well as chunking mechanisms on all models to ensure fair comparison by processing all input frames at once.

Results in Table \ref{tab:efficiency} show that our student model is nearly $5\times$ faster than the teacher model at $N=128$ views. This divergence grows as the number of views increases. The difference in inference speed is more pronounced with VGGT where Distill3R is nearly $6\times$ faster by  $N=64$ views. 

\begin{table*}[t]
\centering
\caption{System efficiency comparison on RTX 6000 Ada. Performance metrics for varying view counts ($N$). Time is measured in seconds (s) and Peak Memory in GB. OOM denotes Out Of Memory. All methods except Fast3R are evaluated at $378 \times 518$ resolution; Fast3R uses $384\times 512$. Bold and underline indicate the best and second-best results, respectively.}
\label{tab:efficiency}
\resizebox{\textwidth}{!}{
\begin{tabular}{lcccccccccc}
\toprule
& \multicolumn{2}{c}{\textbf{$N=12$}} & \multicolumn{2}{c}{\textbf{$N=32$}} & \multicolumn{2}{c}{\textbf{$N=64$}} & \multicolumn{2}{c}{\textbf{$N=96$}} & \multicolumn{2}{c}{\textbf{$N=128$}} \\
\cmidrule(lr){2-3} \cmidrule(lr){4-5} \cmidrule(lr){6-7} \cmidrule(lr){8-9} \cmidrule(lr){10-11}
\textbf{Method} & Time (s) & Mem (GB) & Time (s) & Mem (GB) & Time (s) & Mem (GB) & Time (s) & Mem (GB) & Time (s) & Mem (GB) \\
\midrule
Fast3R (Teacher) \cite{yang2025fast3r} & \underline{0.32} & \underline{6.86} & \underline{1.14} & \underline{12.11} & \underline{3.26} & \underline{21.11} & \underline{6.35} & \underline{32.36} & \underline{10.11} & \underline{44.36} \\
VGGT \cite{wang2025vggt} & 0.59 & 15.28 & 2.28 & 33.98 & 6.40 & 38.41 & OOM & OOM & OOM & OOM \\
\midrule
\textbf{Distill3R (Ours)} & \textbf{0.13} & \textbf{4.05} & \textbf{0.41} & \textbf{9.97} & \textbf{1.02} & \textbf{21.80} & \textbf{1.78} & \textbf{28.69} & \textbf{2.69} & \textbf{31.90} \\
\bottomrule
\end{tabular}
}
\end{table*}

\subsection{3D Reconstruction}
\label{subsec:recon}

We evaluate the geometric reconstruction quality of our distilled student model on unseen scene-level (7-Scenes \cite{shotton2013scene}) and object-level (DTU \cite{aanaes2016large}) benchmarks. To isolate the student's learned geometric reasoning from rotational bottlenecks in global alignment, we perform a per-view evaluation. Fast3R also evaluates reconstruction using the local head while using ICP for global alignment \cite{yang2025fast3r}.

As reported in Table \ref{tab:reconstruction}, the student model exhibits a characterized trade-off between absolute precision and metric grounding. While the teacher maintains higher fine-grained regression accuracy, our $72~\mathrm{M}$ student demonstrates a significantly superior adherence to absolute metric scale across both indoor and extreme out-of-distribution (OOD) object-centric scenes.

\begin{figure*}[t!]
\centering
\begin{tabular}{ccc}
    \includegraphics[width=0.31\textwidth]{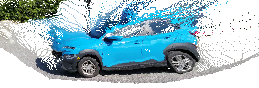} & 
    \includegraphics[width=0.31\textwidth]{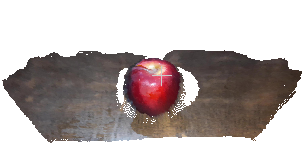} & 
    \includegraphics[width=0.31\textwidth]{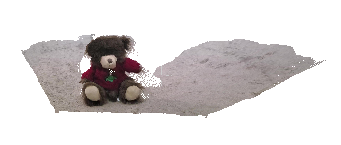} \\
    (a) Car & (b) Apple & (c) Teddy Bear
\end{tabular}
\caption{Object reconstruction (CO3D-v2). Qualitative results for the Distill3R student on object-centric scenes. }
\label{fig:co3d_qual}
\end{figure*}

On the 7-Scenes benchmark, the student achieves an average scale factor of 1.62, closely approaching true metric scale (1.0). In contrast, the  $650~\mathrm{M}$ teacher exhibits an average scale factor of 3.54, requiring heavy post-hoc scaling to match real-world dimensions. These results suggest that the distillation process regularizes the student toward a more grounded metric prior. For robotics applications, such as navigation initialization and reactive mapping, this metric stability is often more valuable than high-frequency textural precision.
While quantitative metrics saturate early, qualitative analysis (Fig. \ref{fig:co3d_qual}) reveals significant improvements in structural coherence and surface smoothness as training progresses to 60 epochs. As observed in Table \ref{tab:ablation_loss}, the primary quantitative shift during extended training is a slight degradation in the completeness metric, while accuracy and metric scale continue to improve.

\begin{figure}[t]
\centering
\begin{tabular}{cc}
\textbf{Fast3R} & \textbf{Distill3R} \\
\includegraphics[width=0.45\columnwidth, height=0.20\columnwidth]{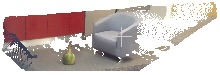} &
\includegraphics[width=0.45\columnwidth]{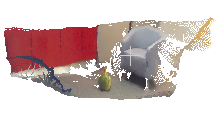} \\
\includegraphics[width=0.45\columnwidth, height=0.20\columnwidth]{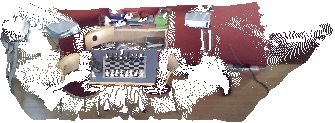} &
\includegraphics[width=0.45\columnwidth]{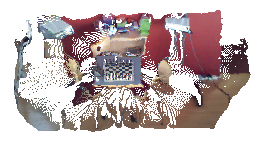} \\
\includegraphics[width=0.45\columnwidth]{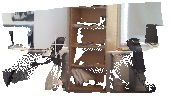} &
\includegraphics[width=0.45\columnwidth]{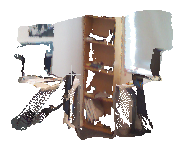}
\end{tabular}
\caption{\textbf{Qualitative comparison on 7-Scenes (OOD).} Side-by-side reconstructions showing that Distill3R (right) preserves the global topology of the teacher (left).}
\label{fig:7scenes_qual}
\end{figure}

\begin{table}[t]
\centering
\caption{Quantitative Reconstruction Results. Per-view evaluation. Metrics are Median Distance (Acc/Comp) scaled by $100\times$. Scale factor measures adherence to true metric scale ($1.0$ is perfect). Bold indicates the best results. Note: 7-Scenes ground truth is in meters, while DTU is in millimeters, accounting for the magnitude difference in scale factors.}
\label{tab:reconstruction}
\resizebox{\columnwidth}{!}{
\begin{tabular}{l ccc ccc}
\toprule
& \multicolumn{3}{c}{\textbf{7-Scenes (OOD)}} & \multicolumn{3}{c}{\textbf{DTU (OOD)}} \\ \cmidrule(lr){2-4} \cmidrule(lr){5-7}
\textbf{Method} & Acc $\downarrow$ & Comp $\downarrow$ & Scale $\downarrow$ & Acc $\downarrow$ & Comp $\downarrow$ & Scale $\downarrow$ \\
\midrule
Fast3R (Teacher) \cite{yang2025fast3r} & \textbf{2.45} & \textbf{2.97} & 3.54 & \textbf{0.31} & \textbf{0.41} & 942.49 \\
\midrule
\textbf{Distill3R (Ours)} & 8.76 & 8.88 & \textbf{1.62} & 1.16 & 1.77 & \textbf{394.59} \\
\bottomrule
\end{tabular}
}
\end{table}

\section{Ablations}
\label{sec:ablations}

\subsection{Encoder}

The first stage is a 2D feature encoder that extracts patch-level embeddings from each input image $I_k$. We selected \textit{DUNE ViT-S} \cite{sariyildiz2025dune} ($21~\mathrm{M}$ parameters) as our backbone. Preliminary ablations showed that DINOv3 ViT-S \cite{simeoni2025dinov3} yielded significantly poorer 3D reconstruction quality. While CroCo-B \cite{leroy2024mast3r} is a strong baseline, the lack of a publicly available ViT-Small variant would necessitate complex architectural modifications and bespoke weight-copying to fit our student's $d=384$ design. DUNE's heterogeneous co-distillation acts as a form of knowledge synthesis, providing a holistic scene representation that is better suited for dense 3D regression than single-modality teachers \cite{sariyildiz2025dune}.

\subsection{Loss Terms and Training Without Distillation}

To validate our distillation objective, we ablated the specific components that leverage teacher uncertainty. These data show that removing these terms degrades the quality of the reconstruction. 

Furthermore, to validate the distillation pipeline, we simulate training from scratch by removing the direct confidence supervision and replacing the teacher confidence weighting on the geometric loss with the student's own confidence. This configuration removes the relational knowledge inherent in distillation, mimicking the standard objective used in models like Fast3R. By treating the teacher's output point maps as pseudo-ground truth, we avoid the computational cost of global Structure-from-Motion (SfM) being used on our entire training dataset while demonstrating the necessity of distillation. This is supported by the fact that the teacher’s point maps are nearly indistinguishable from traditional SfM results in terms of structural consistency. As shown in Table \ref{tab:ablation_loss}, the absence of distillation signals results in a 9.6\% reduction in accuracy and a substantial 35.7\% degradation in completion quality.

\begin{table}[t]
\centering
\caption{Loss Term Ablation. Evaluated on DTU (OOD). These data show that geometric coordinate transfer alone (Labels-Only) is insufficient without teacher uncertainty signals. Bold and underline indicate the best and second-best results, respectively.} 
\label{tab:ablation_loss}
\resizebox{\columnwidth}{!}{
\begin{tabular}{lccc}
\toprule
Configuration & Acc $\downarrow$ & Comp $\downarrow$ & Scale $\downarrow$ \\
\midrule
Labels-Only (No Weighting/Conf) & \underline{1.37} & 2.09 & \underline{409.05} \\
w/o Weighting $w_{k,ij}$ & 1.42 & \underline{1.88} & 467.85 \\
\midrule
Distill3R (Full) & \textbf{1.25} & \textbf{1.54} & \textbf{380.30} \\
\bottomrule
\end{tabular}
}
\end{table}

\subsection{Discussion: A Roadmap for Efficient 3D Vision}
\label{sec:limitations}

Distill3R serves as a foundational entry point for democratized research into 3D foundation models. By providing a pipeline that can be trained on a single workstation, we lay the groundwork for future improvements in neural reconstruction.

\textit{The Pose Estimation Challenge.} A primary structural limitation of the current Distill3R iteration is rotational drift in global pose estimation. The student achieves accurate translation in the global reference frame but fails to capture the required rotation to properly align views. Because the Fast3R teacher relies on ICP-based alignment to ground-truth poses during its own training, the student receives geometry that is already anchored to a reference frame without explicitly learning the rotation parameters. Using ground-truth poses combined with procedural alignment at train time would impose a significant bottleneck on dataset construction and cache generation, reducing the accessibility of the pipeline for the research community. 
Addressing this limitation by directly supervising camera parameters would require a substantially different training formulation and supervision signal, shifting the focus from geometry distillation toward explicit pose regression. In this context, permutation-equivariant models such as $\mathrm{\pi}^3$ \cite{wang2026pi3} represent a promising direction for future work, as they are designed to infer camera parameters without reliance on a fixed reference frame. Integrating such models into the Distill3R pipeline would enable end-to-end pose supervision, but would also entail a non-trivial expansion of the training pipeline and evaluation scope beyond the goals of the current work.

\textit{Enhancing Robustness and Metrics.} To further bridge the quantitative gap identified in our benchmarks, we identify several promising directions:
\begin{itemize}
    \item \textit{Data Augmentation:} Implementing aspect-ratio and scale-invariant augmentations accompanied by a matching logic in the teacher cache could improve robustness to varying sensor characteristics.
    \item \textit{Geometric Priors:} Integrating a planar loss (Manhattan-world prior) could significantly improve the metrics on indoor benchmarks like 7-Scenes where warping is present.
\end{itemize}

\textit{Specialized Knowledge Synthesis.} The Distill3R pipeline provides a template for domain-specific specialization. By decoupling teacher inference, researchers can efficiently train compact student models on proprietary or niche datasets such as medical imaging or specific industrial environments, where a general-purpose foundation model may be too computationally expensive to deploy. This enables the creation of highly specialized, low-latency models tailored for specific sub-tasks without requiring the massive infrastructure used for initial foundation model training.

\section{Conclusion}
\label{sec:conlusion}

In this work, we introduced Distill3R, a novel distillation framework designed to close the gap between heavyweight 3D foundation models and resource-constrained research and deployment environments. We demonstrate that the structural intelligence of $650~\mathrm{M}$-parameter models can be distilled into a student an order of magnitude smaller, achieving significantly superior adherence to absolute metric scale despite a drop in fine-grained precision. By reducing the training infrastructure requirements to a single workstation and providing a $5\times$ speedup at inference time, we have laid the groundwork for future iterations to address the current global pose estimation bottlenecks by providing a modular pipeline that can integrate next-generation permutation-equivariant teachers. Distill3R serves as a proof-of-concept that high-capacity 3D vision can be democratized. By shifting the focus from over-parameterization to efficient distillation and knowledge synthesis, the community can accelerate the deployment of these models onto mobile robots and embedded systems. Additionally, Distill3R provides an accessible research baseline that enables laboratories without large-scale compute to train and specialize 3D models on their own domain-specific data at minimal cost. 
\section*{Acknowledgment}
\label{sec:ack}

This research was undertaken, in part, based on support from the Natural Sciences and Engineering Research Council of Canada Grant RGPIN-2021-03479 (NSERC DG) and through the Natural Sciences and Engineering Research Council of Canada CGS-M scholarship (NSERC CGS-M).

\bibliographystyle{IEEEtran}
\bibliography{master.bib} 

@STRING{cvpr	= {Proceedings of the IEEE/CVF Conference on Computer Vision
		  and Pattern Recognition} }

@STRING{cvprw	= {Proceedings of the IEEE/CVF Conference on Computer Vision
		  and Pattern Recognition Workshops} }

@STRING{eccv	= {Proceedings of the European Conference on Computer
		  Vision} }

@STRING{iccv	= {Proceedings of the IEEE/CVF International Conference on
		  Computer Vision} }

@STRING{iclr	= {Proceedings of the International Conference on Learning
		  Representations} }

@STRING{icml	= {Proceedings of the International Conference on Machine
		  Learning} }

@STRING{ijcv	= {International Journal of Computer Vision} }

@STRING{neurips	= {Advances in Neural Information Processing Systems} }

@STRING{siggraph= {Proceedings of ACM SIGGRAPH} }

@STRING{threedv	= {Proceedings of the International Conference on 3D Vision} }

@Article{	  simeoni2025dinov3,
  author	= {Sim{\'e}oni, Oriane and Vo, Huy V. and Seitzer, Maximilian
		  and Baldassarre, Federico and Oquab, Maxime and Jose, Cijo
		  and Khalidov, Vasil and Szafraniec, Marc and Yi, Seungeun
		  and Ramamonjisoa, Micha{\"e}l and others},
  title		= {{DINOv3}},
  journal	= {arXiv preprint arXiv:2508.10104},
  year		= {2025}
}

@InProceedings{	  wang2025vggt,
  title		= {{VGGT}: Visual geometry grounded transformer},
  author	= {Wang, Jianyuan and Chen, Minghao and Karaev, Nikita and
		  Vedaldi, Andrea and Rupprecht, Christian and Novotny,
		  David},
  booktitle	= cvpr,
  pages		= {5294--5306},
  year		= {2025}
}

@InProceedings{	  wang2026pi3,
  author	= {Wang, Yifan and Zhou, Jianjun and Zhu, Haoyi and Chang,
		  Wenzheng and Zhou, Yang and Li, Zizun and Chen, Junyi and
		  Pang, Jiangmiao and Shen, Chunhua and He, Tong},
  title		= {{$\pi^3$}: Permutation-Equivariant Visual Geometry
		  Learning},
  booktitle	= iclr,
  year		= {2026}
}

@InProceedings{	  agarwal2009building,
  title		= {Building {Rome} in a day},
  author	= {Agarwal, Sameer and Snavely, Noah and Simon, Ian and
		  Seitz, Steven M. and Szeliski, Richard},
  booktitle	= iccv,
  pages		= {72--79},
  year		= {2009}
}

@InProceedings{	  baruch2021arkitscenes,
  title		= {{ARKitScenes}: A diverse real-world dataset for {3D}
		  indoor scene understanding using mobile {RGB-D} data},
  author	= {Baruch, Gilad and Chen, Zhuoyuan and Dehghan, Afshin and Dimry, Tal and Feigin, Yuri and Fu, Peter and Gebauer, Thomas and Joffe, Brandon and Kurz, Daniel and Schwartz, Arik and others},
  booktitle	= neurips,
  year		= {2021}
}

@InProceedings{	  dai2017scannet,
  title		= {{ScanNet}: Richly-annotated {3D} reconstructions of indoor
		  scenes},
  author	= {Dai, Angela and others},
  booktitle	= cvpr,
  pages		= {5828--5839},
  year		= {2017}
}

@InProceedings{	  deitke2023objaverse,
  title		= {{Objaverse}: A universe of annotated {3D} objects},
  author	= {Deitke, Matt and others},
  booktitle	= cvpr,
  pages		= {13142--13153},
  year		= {2023}
}

@InProceedings{	  detone2018superpoint,
  title		= {{SuperPoint}: Self-supervised interest point detection and
		  description},
  author	= {DeTone, Daniel and Malisiewicz, Tomasz and Rabinovich,
		  Andrew},
  booktitle	= cvprw,
  pages		= {224--236},
  year		= {2018}
}

@InProceedings{	  ding2022kdmvs,
  title		= {{KD-MVS}: Knowledge distillation based self-supervised
		  learning for multi-view stereo},
  author	= {Ding, Yikang and Zhu, Qingtian and Liu, Xiangyue and Yuan,
		  Wentao and Zhang, Haotian and Zhang, Chi},
  booktitle	= eccv,
  pages		= {630--646},
  year		= {2022},
  organization	= {Springer}
}

@InProceedings{	  dosovitskiy2021image,
  title		= {An image is worth 16x16 words: Transformers for image
		  recognition at scale},
  author	= {Dosovitskiy, Alexey and others},
  booktitle	= iclr,
  year		= {2021}
}

@InProceedings{	  dusmanu2019d2net,
  title		= {{D2-Net}: A trainable {CNN} for joint description and
		  detection of local features},
  author	= {Dusmanu, Mihai and others},
  booktitle	= cvpr,
  pages		= {8092--8101},
  year		= {2019}
}

@Article{	  dutt2024multiview,
  title		= {Multi-View {3D} Reconstruction using Knowledge
		  Distillation},
  author	= {Dutt, Aditya and Lunawat, Ishikaa and Kaur, Manpreet},
  journal	= {arXiv preprint arXiv:2412.02039},
  year		= {2024}
}

@InProceedings{	  edstedt2023dkm,
  title		= {{DKM}: Dense kernelized feature matching for geometry
		  estimation},
  author	= {Edstedt, Johan and Athanasiadis, Ioannis and
		  Wadenb{\"{a}}ck, M{\r{a}}rten and Felsberg, Michael},
  booktitle	= cvpr,
  year		= {2023},
  pages		= {17765--17775}
}

@InProceedings{	  edstedt2024roma,
  title		= {{RoMa}: Robust dense feature matching},
  author	= {Edstedt, Johan and Sun, Qiyu and B{\"{o}}kman, Georg and
		  Wadenb{\"{a}}ck, M{\r{a}}rten and Felsberg, Michael},
  booktitle	= cvpr,
  pages		= {19790--19800},
  year		= {2024}
}

@InProceedings{	  fastvggt_arxiv,
  title		= {{FastVGGT}: Fast visual geometry transformer},
  author	= {Shen, You and Zhang, Zhipeng and Qu, Yansong and Zheng,
		  Xiawu and Ji, Jiayi and Zhang, Shengchuan and Cao,
		  Liujuan},
  booktitle	= iclr,
  year		= {2026}
}

@Article{	  gou2020knowledge,
  title		= {Knowledge distillation: A survey},
  author	= {Gou, Jianping and Yu, Baosheng and Maybank, Stephen J. and
		  Tao, Dacheng},
  journal	= ijcv,
  year		= {2021},
  volume	= {129},
  number	= {6},
  pages		= {1789--1819}
}

@InProceedings{	  gu2020cascade,
  title		= {Cascade cost volume for high-resolution multi-view stereo
		  and stereo matching},
  author	= {Gu, Xiaodong and others},
  booktitle	= cvpr,
  pages		= {2495--2504},
  year		= {2020}
}

@Book{		  hartley2004multiple,
  title		= {Multiple View Geometry in Computer Vision},
  author	= {Hartley, Richard and Zisserman, Andrew},
  publisher	= {Cambridge University Press},
  year		= {2004},
  edition	= {2nd}
}

@Article{	  hinton2015distillation,
  title		= {Distilling the knowledge in a neural network},
  author	= {Hinton, Geoffrey E. and Vinyals, Oriol and Dean, Jeffrey},
  journal	= {arXiv preprint arXiv:1503.02531},
  year		= {2015}
}

@InProceedings{	  kendall2017what,
  title		= {What uncertainties do we need in {B}ayesian deep learning
		  for computer vision?},
  author	= {Kendall, Alex and Gal, Yarin},
  booktitle	= neurips,
  volume	= {30},
  year		= {2017}
}

@Article{	  lepetit2009epnp,
  title		= {{EPnP}: An accurate {$O(n)$} solution to the {PnP}
		  problem},
  author	= {Lepetit, Vincent and Moreno-Noguer, Francesc and Fua,
		  Pascal},
  journal	= {International Journal of Computer Vision},
  volume	= {81},
  pages		= {155--166},
  year		= {2009},
  publisher	= {Springer}
}

@InProceedings{	  leroy2024mast3r,
  title		= {Grounding Image Matching in {3D} with {MASt3R}},
  author	= {Leroy, Vincent and Cabon, Yohann and Revaud, J\'er\^ome},
  booktitle	= eccv,
  year		= {2024},
  pages		= {71--91}
}

@InProceedings{	  li2018megadepth,
  title		= {{MegaDepth}: Learning single-view depth prediction from
		  internet photos},
  author	= {Li, Zhengqi and Snavely, Noah},
  booktitle	= cvpr,
  pages		= {2041--2050},
  year		= {2018}
}

@InProceedings{	  lindenberger2023lightglue_iccv,
  title		= {{LightGlue}: Local feature matching at light speed},
  author	= {Lindenberger, Philipp and Sarlin, Paul-Edouard and
		  Pollefeys, Marc},
  booktitle	= iccv,
  year		= {2023},
  pages		= {17627-17638}
}

@Article{	  ozyesil2017survey,
  title		= {A survey of structure from motion},
  author	= {{\"O}zye{\c{s}}il, Onur and Voroninski, Vladislav and
		  Basri, Ronen and Singer, Amit},
  journal	= {Acta Numerica},
  volume	= {26},
  pages		= {305--364},
  year		= {2017}
}

@InProceedings{	  quantvggt_arxiv,
  title		= {Quantized visual geometry grounded transformer},
  author	= {Feng, Weilun and others},
  booktitle	= iclr,
  year		= {2026}
}

@InProceedings{	  reizenstein2021co3d,
  title		= {Common objects in {3D}: Large-scale learning and
		  evaluation of real-life {3D} category reconstruction},
  author	= {Reizenstein, Jeremy and others},
  booktitle	= iccv,
  year		= {2021},
  pages		= {10901--10911}
}

@InProceedings{	  sariyildiz2025dune,
  title		= {{DUNE}: Distilling a universal encoder from heterogeneous
		  {2D} and {3D} teachers},
  author	= {Sar{\i}y{\i}ld{\i}z, M. B{\"u}lent and others},
  booktitle	= cvpr,
  year		= {2025}
}

@InProceedings{	  sarlin2020superglue,
  title		= {{SuperGlue}: Learning feature matching with graph neural
		  networks},
  author	= {Sarlin, Paul-Edouard and DeTone, Daniel and Malisiewicz,
		  Tomasz and Rabinovich, Andrew},
  booktitle	= cvpr,
  pages		= {4938--4947},
  year		= {2020}
}

@InProceedings{	  savva2019habitat,
  title		= {Habitat: A platform for embodied {AI} research},
  author	= {Savva, Manolis and others},
  booktitle	= iccv,
  year		= {2019},
  pages		= {9339--9347}
}

@InProceedings{	  schonberger2016sfm,
  title		= {Structure-from-motion revisited},
  author	= {Sch{\"{o}}nberger, Johannes Lutz and Frahm, Jan-Michael},
  booktitle	= cvpr,
  year		= {2016},
  pages		= {4104--4113}
}

@InProceedings{	  snavely2006photo,
  title		= {Photo tourism: Exploring photo collections in {3D}},
  author	= {Snavely, Noah and Seitz, Steven M. and Szeliski, Richard},
  booktitle	= siggraph,
  pages		= {835--846},
  year		= {2006}
}

@Article{	  song2023msdpt,
  title		= {Knowledge distillation of multi-scale dense prediction
		  transformer for self-supervised depth estimation},
  author	= {Song, Jimin and Lee, Sang Jun},
  journal	= {Sci. Rep.},
  volume	= {13},
  number	= {1},
  pages		= {18939},
  year		= {2023}
}

@InProceedings{	  sun2021loftr,
  title		= {{LoFTR}: Detector-free local feature matching with
		  transformers},
  author	= {Sun, Jiaming and Shen, Zehong and Wang, Yuang and Bao,
		  Hujun and Zhou, Xiaowei},
  booktitle	= cvpr,
  pages		= {8922--8931},
  year		= {2021}
}

@InProceedings{	  tang2018ba,
  title		= {{BA-Net}: Dense bundle adjustment networks},
  author	= {Tang, Chengzhou and Tan, Ping},
  booktitle	= iclr,
  year		= {2019}
}

@InProceedings{	  tang2024mvdust3r,
  title		= {{MV-DUSt3R+}: Single-stage scene reconstruction from
		  sparse views in 2 seconds},
  author	= {Tang, Zhenggang and Fan, Yuchen and Wang, Dilin and Xu,
		  Hongyu and Ranjan, Rakesh and Schwing, Alexander and Yan,
		  Zhicheng},
  booktitle	= cvpr,
  pages		= {5283--5293},
  year		= {2025}
}

@InProceedings{	  teed2018deepv2d,
  title		= {{DeepV2D}: Video to depth with differentiable structure
		  from motion},
  author	= {Teed, Zachary and Deng, Jia},
  booktitle	= iclr,
  year		= {2020}
}

@InProceedings{	  touvron2021training,
  title		= {Training data-efficient image transformers \& distillation
		  through attention},
  author	= {Touvron, Hugo and others},
  booktitle	= icml,
  pages		= {10347--10357},
  year		= {2021}
}

@InProceedings{	  tyszkiewicz2020disk,
  title		= {{DISK}: Learning local features with policy gradient},
  author	= {Tyszkiewicz, Micha{\l} and Fua, Pascal and Trulls, Eduard},
  booktitle	= neurips,
  volume	= {33},
  pages		= {14254--14265},
  year		= {2020}
}

@InProceedings{	  wang2023posediffusion,
  title		= {{PoseDiffusion}: Solving pose estimation via
		  diffusion-aided bundle adjustment},
  author	= {Wang, Jianyuan and Rupprecht, Christian and Novotny,
		  David},
  booktitle	= iccv,
  pages		= {9773--9783},
  year		= {2023}
}

@InProceedings{	  wang2024dust3r,
  title		= {{DUSt3R}: Geometric {3D} vision made easy},
  author	= {Wang, Shuzhe and Leroy, Vincent and Cabon, Yohann and
		  Chidlovskii, Boris and Revaud, Jerome},
  booktitle	= cvpr,
  year		= {2024},
  pages		= {20697--20709}
}

@InProceedings{	  wang2024vggsfm,
  title		= {{VGGSfM}: Visual geometry grounded deep structure from
		  motion},
  author	= {Wang, Jianyuan and Karaev, Nikita and Rupprecht, Christian
		  and Novotny, David},
  booktitle	= cvpr,
  year		= {2024},
  pages		= {21686--21697}
}

@InProceedings{	  wei2020deepsfm,
  title		= {{DeepSFM}: Structure from motion via deep bundle
		  adjustment},
  author	= {Wei, Xingkui and Zhang, Yinda and Li, Zhuwen and Fu,
		  Yanwei and Xue, Xiangyang},
  booktitle	= eccv,
  pages		= {230--247},
  year		= {2020},
  organization	= {Springer}
}

@InProceedings{	  wu2013towards,
  title		= {Towards linear-time incremental structure from motion},
  author	= {Wu, Changchang},
  booktitle	= threedv,
  pages		= {127--134},
  year		= {2013},
  organization	= {IEEE}
}

@InProceedings{	  wu2022tinyvit,
  title		= {{TinyViT}: Fast pretraining distillation for small vision
		  transformers},
  author	= {Wu, Kan and others},
  booktitle	= eccv,
  year		= {2022}
}

@InProceedings{	  wu2023adudepth,
  title		= {{ADU-Depth}: Attention-based distillation with uncertainty
		  modeling for depth estimation},
  author	= {Wu, Zizhang and Li, Zhuozheng and Fan, Zhi-Gang and Wu,
		  Yunzhe and Wang, Xiaoquan and Tang, Rui and Pu, Jian},
  booktitle	= {Conference on Robot Learning},
  pages		= {3167--3179},
  year		= {2023},
  organization	= {PMLR}
}

@InProceedings{	  yang2025fast3r,
  title		= {{Fast3R}: Towards {3D} reconstruction of 1000+ images in
		  one forward pass},
  author	= {Yang, Jianing and others},
  booktitle	= cvpr,
  year		= {2025},
  pages		= {21924-21935}
}

@InProceedings{	  yao2018mvsnet,
  title		= {{MVSNet}: Depth inference for unstructured multi-view
		  stereo},
  author	= {Yao, Yao and Luo, Zixin and Li, Shiwei and Fang, Tian and
		  Quan, Long},
  booktitle	= eccv,
  year		= {2018}
}

@InProceedings{	  yao2020blendedmvs,
  title		= {{BlendedMVS}: A large-scale dataset for generalized
		  multi-view stereo networks},
  author	= {Yao, Yao and others},
  booktitle	= cvpr,
  pages		= {1790--1799},
  year		= {2020}
}

@InProceedings{	  yeshwanth2023scannetpp,
  title		= {{ScanNet++}: A high-fidelity dataset of {3D} indoor
		  scenes},
  author	= {Yeshwanth, Chandan and Liu, Yueh-Cheng and Nie{\ss}ner,
		  Matthias and Dai, Angela},
  booktitle	= iccv,
  year		= {2023},
  pages		= {12-22}
}

@InProceedings{	  yi2016lift,
  title		= {{LIFT}: Learned invariant feature transform},
  author	= {Yi, Kwang Moo and Trulls, Eduard and Lepetit, Vincent and
		  Fua, Pascal},
  booktitle	= eccv,
  pages		= {467--483},
  year		= {2016}
}

@InProceedings{	  zhang2025monst3r,
  title		= {{MonST3R}: A Simple Approach for Estimating Geometry in
		  the Presence of Motion},
  author	= {Zhang, Junyi and Herrmann, Charles and Hur, Junhwa and
		  Jampani, Varun and Darrell, Trevor and Cole, Forrester and
		  Sun, Deqing and Yang, Ming-Hsuan},
  booktitle	= iclr,
  year		= {2025}
}

@InProceedings{	  wang2025spann3r,
  title		= {{3D} reconstruction with spatial memory},
  author	= {Wang, Hengyi and Agapito, Lourdes},
  booktitle	= threedv,
  pages		= {78--89},
  year		= {2025},
  organization	= {IEEE}
}

@InProceedings{	  zhang2025flare,
  title		= {{FLARE}: Feed-forward geometry, appearance and camera
		  estimation from uncalibrated sparse views},
  author	= {Zhang, Shangzhan and others},
  booktitle	= cvpr,
  year		= {2025},
  pages		= {21936--21947}
}

@Article{	  zhao2023aliked,
  title		= {{ALIKED}: A lighter keypoint and descriptor extraction
		  network via deformable transformation},
  author	= {Zhao, Xiaoming and others},
  journal	= {IEEE Trans. Instrum. Meas.},
  volume	= {72},
  pages		= {1--16},
  year		= {2023}
}

@Article{	  kaplan2020scaling,
  title		= {Scaling laws for neural language models},
  author	= {Kaplan, Jared and others},
  journal	= {arXiv preprint arXiv:2001.08361},
  year		= {2020}
}

@Article{	  hoffmann2022training,
  title		= {Training compute-optimal large language models},
  author	= {Hoffmann, Jordan and others},
  journal	= {arXiv preprint arXiv:2203.15556},
  year		= {2022}
}

@InProceedings{	  cabon2025must3r,
  title		= {{MUSt3R}: Multi-view network for stereo {3D}
		  reconstruction},
  author	= {Cabon, Yohann and others},
  booktitle	= cvpr,
  year		= {2025}
}

@InProceedings{	  jang2025pow3r,
  title		= {{Pow3R}: Empowering unconstrained {3D} reconstruction with
		  camera and scene priors},
  author	= {Jang, Wonbong and Weinzaepfel, Philippe and Leroy, Vincent
		  and Agapito, Lourdes and Revaud, Jerome},
  booktitle	= cvpr,
  year		= {2025}
}

@Article{	  lowe2004distinctive,
  title		= {Distinctive image features from scale-invariant
		  keypoints},
  author	= {Lowe, David G.},
  journal	= ijcv,
  year		= {2004},
  volume	= {60},
  number	= {2},
  pages		= {91--110}
}

@InProceedings{	  kendall2015posenet,
  title		= {{PoseNet}: A convolutional network for real-time {6-DOF}
		  camera relocalization},
  author	= {Kendall, Alex and Grimes, Matthew and Cipolla, Roberto},
  booktitle	= iccv,
  pages		= {2938--2946},
  year		= {2015}
}

@InProceedings{	  shotton2013scene,
  title		= {Scene coordinate regression forests for camera
		  relocalization in {RGB-D} images},
  author	= {Shotton, Jamie and others},
  booktitle	= cvpr,
  pages		= {2930--2937},
  year		= {2013}
}

@Article{	  aanaes2016large,
  title		= {Large-scale data for multiple-view stereopsis},
  author	= {Aan{\ae}s, Henrik and others},
  journal	= ijcv,
  volume	= {120},
  pages		= {153--168},
  year		= {2016}
}

@InProceedings{	  sun2025avggtrethinkingglobalattention,
  title		= {{AVGGT}: Rethinking global attention for accelerating
		  {VGGT}},
  author	= {Sun, Xianbing and others},
  booktitle	= cvpr,
  year		= {2026}
}

@InProceedings{	  wang2025fastervggtblocksparseglobal,
  title		= {Block-sparse global attention for efficient multi-view
		  geometry transformers},
  author	= {Wang, Chung-Shien Brian and Schmidt, Christian and
		  Piekenbrinck, Jens and Leibe, Bastian},
  booktitle	= cvpr,
  year		= {2026}
}

@Article{	  stary2025understandingmultiviewtransformers,
  title		= {Understanding multi-view transformers},
  author	= {Stary, Michal and Gaubil, Julien and Tewari, Ayush and
		  Sitzmann, Vincent},
  journal	= {arXiv preprint arXiv:2510.24907},
  year		= {2025}
}

@article{achiam2023gpt4,
  title={{GPT-4} technical report},
  author={{OpenAI} and Achiam, Josh and Adler, Steven and Agarwal, Sandhini and Ahmad, Lama and Akkaya, Ilge and Aleman, Florencia Leoni and Almeida, Diogo and Altenschmidt, Janko and Altman, Sam and Anadkat, Shyamal and others},
  journal={arXiv preprint arXiv:2303.08774},
  year={2023}
}

@inproceedings{cheng2022masked,
  title={Masked-attention mask transformer for universal image segmentation},
  author={Cheng, Bowen and Misra, Ishan and Schwing, Alexander G and Kirillov, Alexander and Girdhar, Rohit},
  booktitle=cvpr,
  pages={1280--1289},
  year={2022},
  organization={IEEE}
}

@inproceedings{dao2024flashattention2,
  title={{FlashAttention-2}: Faster attention with better parallelism and work partitioning},
  author={Dao, Tri},
  booktitle=iclr,
  volume={2024},
  pages={35549--35562},
  year={2024}
}

@inproceedings{peebles2023scalable,
  title={Scalable diffusion models with transformers},
  author={Peebles, William and Xie, Saining},
  booktitle=iccv,
  pages={4172--4182},
  year={2023},
  organization={IEEE}
}

@InProceedings{szot2021habitat,
  title={Habitat 2.0: Training home assistants to rearrange their habitat},
  author={Szot, Andrew and Clegg, Alexander and Undersander, Eric and Wijmans, Erik and Zhao, Yili and Turner, John and Maestre, Noah and Mukadam, Mustafa and Chaplot, Devendra Singh and Maksymets, Oleksandr and others},
  booktitle=neurips,
  volume={34},
  pages={251--266},
  year={2021}
}

@inproceedings{wei2021nerfingmvs,
  title={{NerfingMVS}: Guided optimization of neural radiance fields for indoor multi-view stereo},
  author={Wei, Yi and Liu, Shaohui and Rao, Yongming and Zhao, Wang and Lu, Jiwen and Zhou, Jie},
  booktitle=iccv,
  pages={5610--5619},
  year={2021}
}

@InProceedings{yang2024depthv2,
  title={{Depth Anything V2}},
  author={Yang, Lihe and Kang, Bingyi and Huang, Zilong and Zhao, Zhen and Xu, Xiaogang and Feng, Jiashi and Zhao, Hengshuang},
  booktitle=neurips,
  volume={37},
  pages={21875--21911},
  year={2024}
}

@InProceedings{	  loshchilov2016sgdr,
  title		= {{SGDR}: Stochastic gradient descent with warm restarts},
  author	= {Loshchilov, Ilya and Hutter, Frank},
  booktitle	= iclr,
  year		= {2017}
}

@Article{	  davison2007monoslam,
  title		= {{MonoSLAM}: Real-time single camera {SLAM}},
  author	= {Davison, Andrew J and Reid, Ian D and Molton, Nicholas D
		  and Stasse, Olivier},
  journal	= {IEEE Transactions on Pattern Analysis and Machine
		  Intelligence},
  volume	= {29},
  number	= {6},
  pages		= {1052--1067},
  year		= {2007},
  publisher	= {IEEE}
}

@InProceedings{brown2020language,
  title={Language models are few-shot learners},
  author={Brown, Tom and Mann, Benjamin and Ryder, Nick and Subbiah, Melanie and Kaplan, Jared D and Dhariwal, Prafulla and Neelakantan, Arvind and Shyam, Pranav and Sastry, Girish and Askell, Amanda and others},
  booktitle=neurips,
  volume={33},
  pages={1877--1901},
  year={2020}
}

@inproceedings{cui2017hsfm,
  title={{HSfM}: Hybrid structure-from-motion},
  author={Cui, Hainan and Gao, Xiang and Shen, Shuhan and Hu, Zhanyi},
  booktitle=cvpr,
  pages={2393--2402},
  year={2017},
  organization={IEEE}
}

@inproceedings{duisterhof2025mast3r,
  title={{MASt3R-SfM}: a fully-integrated solution for unconstrained structure-from-motion},
  author={Duisterhof, Bardienus Pieter and Zust, Lojze and Weinzaepfel, Philippe and Leroy, Vincent and Cabon, Yohann and Revaud, Jerome},
  booktitle=threedv,
  pages={1--10},
  year={2025},
  organization={IEEE}
}

@inproceedings{radford2021learning,
  title={Learning transferable visual models from natural language supervision},
  author={Radford, Alec and Kim, Jong Wook and Hallacy, Chris and Ramesh, Aditya and Goh, Gabriel and Agarwal, Sandhini and Sastry, Girish and Askell, Amanda and Mishkin, Pamela and Clark, Jack and others},
  booktitle=icml,
  pages={8748--8763},
  year={2021},
  organization={PMLR}
}

@inproceedings{tang2024lgm,
  title={{LGM}: Large multi-view {Gaussian} model for high-resolution {3D} content creation},
  author={Tang, Jiaxiang and Chen, Zhaoxi and Chen, Xiaokang and Wang, Tengfei and Zeng, Gang and Liu, Ziwei},
  booktitle=eccv,
  pages={1--18},
  year={2024},
  organization={Springer}
}

@InProceedings{weinzaepfel2022croco,
  title={{CroCo}: Self-supervised pre-training for {3D} vision tasks by cross-view completion},
  author={Weinzaepfel, Philippe and Leroy, Vincent and Lucas, Thomas and Br{\'e}gier, Romain and Cabon, Yohann and Arora, Vaibhav and Antsfeld, Leonid and Chidlovskii, Boris and Csurka, Gabriela and Revaud, J{\'e}r{\^o}me},
  booktitle=neurips,
  volume={35},
  pages={3502--3516},
  year={2022}
}

@inproceedings{zheng2023pointodyssey,
  title={{PointOdyssey}: A large-scale synthetic dataset for long-term point tracking},
  author={Zheng, Yang and Harley, Adam W and Shen, Bokui and Wetzstein, Gordon and Guibas, Leonidas J},
  booktitle=iccv,
  pages={19798--19808},
  year={2023},
  organization={IEEE}
}

@inproceedings{micikevicius2017mixed,
  title={Mixed precision training},
  author={Micikevicius, Paulius and Narang, Sharan and Alben, Jonah and Diamos, Gregory and Elsen, Erich and Garcia, David and Ginsburg, Boris and Houston, Michael and Kuchaiev, Oleksii and Venkatesh, Ganesh and others},
  booktitle=iclr,
  year={2018}
}

@Article{	  qin2018vins,
  title		= {{VINS-Mono}: A robust and versatile monocular
		  visual-inertial state estimator},
  author	= {Qin, Tong and Li, Peiliang and Shen, Shaojie},
  journal	= {IEEE Transactions on Robotics},
  volume	= {34},
  number	= {4},
  pages		= {1004--1020},
  year		= {2018},
  publisher	= {IEEE}
}

@inproceedings{caron2021emerging,
  title={Emerging properties in self-supervised vision transformers},
  author={Caron, Mathilde and Touvron, Hugo and Misra, Ishan and J{\'e}gou, Herv{\'e} and Mairal, Julien and Bojanowski, Piotr and Joulin, Armand},
  booktitle=iccv,
  pages={9630--9640},
  year={2021},
  organization={IEEE}
}

@InProceedings{dao2022flashattention,
  title={{FlashAttention}: Fast and memory-efficient exact attention with {IO}-awareness},
  author={Dao, Tri and Fu, Dan and Ermon, Stefano and Rudra, Atri and R{\'e}, Christopher},
  booktitle=neurips,
  volume={35},
  pages={16344--16359},
  year={2022}
}

@InProceedings{	  pan2023aria,
  title		= {{Aria Digital Twin}: A new benchmark dataset for egocentric
		  {3D} machine perception},
  author	= {Pan, Xiaqing and Charron, Nicholas and Yang, Yongqian and Peters, Scott and Whelan, Thomas and Kong, Chen and Parkhi, Omkar and Newcombe, Richard and Ren, Yuheng Carl},
  booktitle	= iccv,
  pages={20076--20086},
  year		= {2023}
}

@inproceedings{ranftl2021vision,
  title={Vision transformers for dense prediction},
  author={Ranftl, Ren{\'e} and Bochkovskiy, Alexey and Koltun, Vladlen},
  booktitle=iccv,
  pages={12159--12168},
  year={2021},
  organization={IEEE}
}

@InProceedings{	  teed2021droid,
  title		= {{DROID-SLAM}: Deep visual {SLAM} for monocular, stereo,
		  and {RGB-D} cameras},
  author	= {Teed, Zachary and Deng, Jia},
  booktitle	= neurips,
  volume	= {34},
  pages		= {16558--16569},
  year		= {2021}
}

@InProceedings{	  xia2024rgbd,
  title		= {{RGBD} objects in the wild: Scaling real-world {3D} object
		  learning from {RGB-D} videos},
  author	= {Xia, Hongchi and Fu, Yang and Liu, Sifei and Wang,
		  Xiaolong},
  booktitle	= cvpr,
  pages		= {22378-22389},
  year		= {2024}
}

@InProceedings{	  loshchilov2017decoupled,
  title		= {Decoupled weight decay regularization},
  author	= {Loshchilov, Ilya and Hutter, Frank},
  booktitle	= iclr,
  year		= {2019}
}

@Article{	  hornung2013octomap,
  title		= {{OctoMap}: An efficient probabilistic {3D} mapping framework
		  based on octrees},
  author	= {Hornung, Armin and Wurm, Kai M and Bennewitz, Maren and
		  Stachniss, Cyrill and Burgard, Wolfram},
  journal	= {Autonomous Robots},
  volume	= {34},
  pages		= {189--206},
  year		= {2013},
  publisher	= {Springer}
}

\end{document}